\documentclass[letterpaper, 10 pt, conference]{ieeeconf}

\usepackage{amsmath,amssymb,amsfonts}
\usepackage{graphicx}
\usepackage{xcolor}
\usepackage{wrapfig}
\usepackage{balance}
\usepackage{dsfont}

\IEEEoverridecommandlockouts 
\title{\LARGE \bf
Walk on Spheres for PDE-based Path Planning
}

\author{Rafael I. Cabral Muchacho and Florian T. Pokorny
\thanks{This work was partially supported by the Wallenberg AI, Autonomous Systems and Software Program (WASP) funded by the Knut and Alice Wallenberg Foundation. The authors are with RPL, EECS, KTH Royal Institute of Technology, Stockholm, Sweden { \tt\small \{ricm, fpokorny\}@kth.se}}}

\begin{document}

\maketitle

\begin{abstract}

In this paper, we investigate the Walk on Spheres algorithm (WoS) for motion planning in robotics. WoS is a Monte Carlo method to solve the Dirichlet problem developed in the 50s by Muller \cite{muller1956some} and has recently been repopularized by Sawhney and Crane \cite{sawhney2020monte} who showed its applicability for geometry processing in volumetric domains. This paper provides a first study into the applicability of WoS for robot motion planning in configuration spaces, with potential fields defined as the solution of screened Poisson equations, motivated by \cite{crane2013geodesics, belyaev2015variational}.
The experiments in this paper empirically indicate the method's trivial parallelization, its dimension-independent convergence characteristic of $O(1/N)$ in the number of walks, and a validation experiment on the RR platform.

\end{abstract}

\IEEEpeerreviewmaketitle

\section{Introduction}

Potential fields are a classic tool in path planning and control \cite{khatib1986real}. Potential fields can be used for path planning by following the negative gradient direction of an artificial potential function, that is repulsive to obstacles and attractive to a goal location. Although potential field methods are reactive and intuitive planning strategies, a main drawback is that the potential function might also have local minima, i.e., can lead to convergence at points different from the goal. To alleviate this drawback, navigation functions were introduced by Koditschek and Rimon \cite{koditschek1990robot}, accompanied by potential field methods leveraging harmonic functions as the solution to Laplace and Poisson partial differential equations (PDE) \cite{connolly1990path, christopher1992applications}.

In this paper we introduce and analyze a generalization of the Laplace and Poisson problems -- the screened Poisson equation -- that implicitly defines a range of smooth and local-minima free potential fields for navigation.

In lower dimensions the screened Poisson equation can be posed and solved through numerical methods, by approximating differential operators through finite differences \cite{sethian1999fast, crane2013geodesics, ames2014numerical}. Although the numerical methods work well in lower dimensions these do not scale well to higher dimensional problems, because of the curse of dimensionality, i.e., memory and computational bottlenecks arising from the exponential growth in necessary resources.

In contrast, grid-free methods do not depend on a discretization of the domain, and have proven to successfully overcome the limitations of traditional methods in lower dimensions. A grid-free approach to solve various PDEs is to leverage Physics Informed Neural Networks (PINN) – an optimization based approach to approximate PDE solutions through neural networks \cite{almqvist2021fundamentals, karniadakis2021physics}. Recently, PINNs have been successfully applied to various Boundary Value Problems (BVP) and also in robot motion planning \cite{ni2023progressive, ni2022ntfields}.

Walk-on-Spheres (WoS), also grid-free, is a Monte Carlo method used to compute the solution of various PDEs within a domain subset. The method was introduced by Muller \cite{muller1956some} and has been further developed and 
applied in multiple fields, such as computational geometry and computer graphics \cite{booth1981exact, sawhney2020monte, qi2022bidirectional, sawhney2023walk}. 

The main contribution of this paper is a method for PDE-based path planning, based on solving the screened Poisson equation through the Walk on Spheres algorithm, enabling PDE-based planning in higher dimensional and arbitrarily shaped domains in configuration spaces.

\begin{figure}
    \centering
    \includegraphics[width=\linewidth]{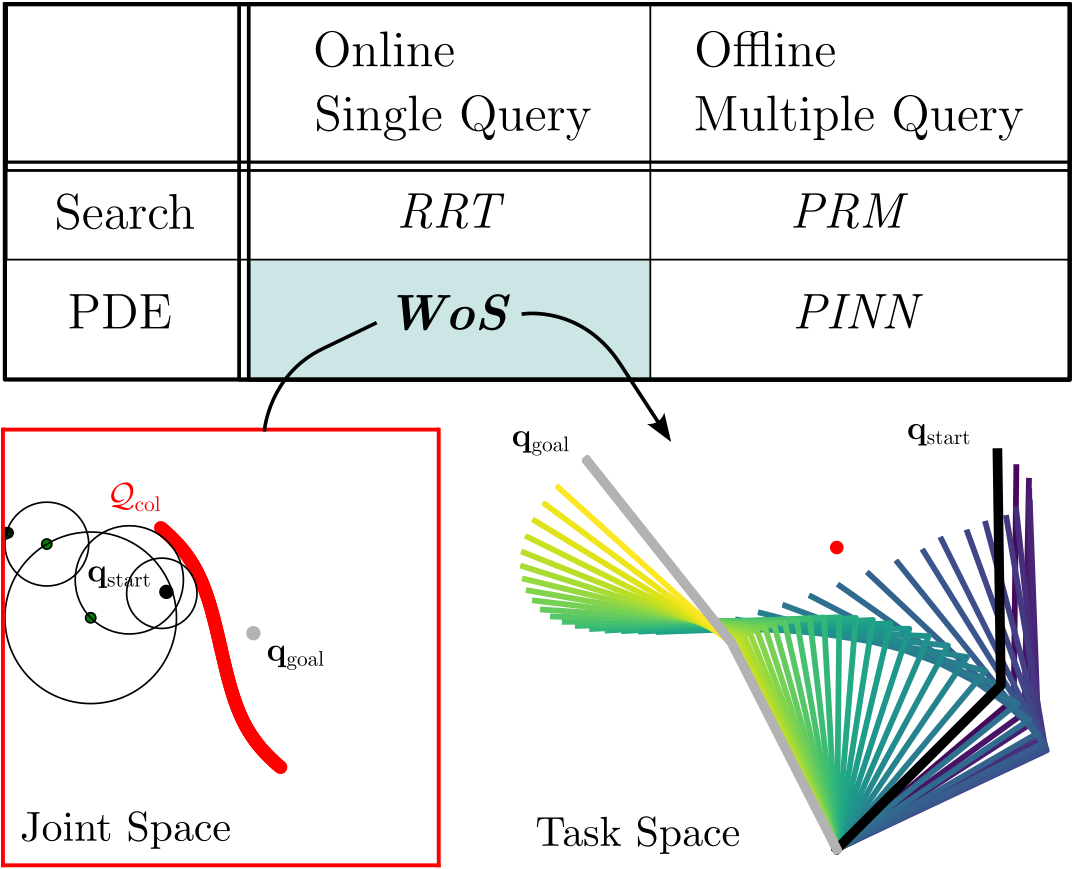}
    \caption{The table shows a non-exhaustive categorization of path planning  strategies applicable in high-dimensional domains. The main contribution of this paper is the introduction of the method in the bottom left corner to robotics, i.e., to compute the solution of PDEs for path planning without prior offline computations leveraging the Walk on Spheres algorithm. The bottom right plot shows the example resulting path in task space, computed using the WoS method.}
    \label{fig:main}
\end{figure}

Given that the WoS algorithm is trivially parallelizable, the empirical trend of exponentially growing GPU computation availability suggests that WoS methods will become increasingly relevant for planning applications.

A robotic application in the realm of holonomic path planning that is enabled by WoS, is the online computation of the gradient ascent direction on a navigation function landscape in higher dimensions ($>3$). In this paper, we focus on this application to empirically evaluate the WoS method and its mentioned properties in a path planning environment.

\section{Related Work}

To understand how the WoS relates to other path planning methods and tools, in this section we describe and extend the categorization table presented in Figure \ref{fig:main}. As described in the previous section and as it is well known in the path planning community, algorithms and methods based on finite discretizations of continuous domains do not scale well into higher dimensions \cite{crane2013geodesics, sethian1999fast, ames2014numerical}. For this reason, we focus only on sampling based methods that have proven their success in higher dimensional applications.

In the table (Fig. \ref{fig:main}) we categorize methods into Search or PDE based, and Online-Single Query or Offline-Multiple Query methods. 
To distinguish between online-single query methods and offline-multiple query methods, the main difference is that the former does not require prior computation and only the path from a specific point to a given goal is computed. Multiple query methods can be advantageous when multiple path computations are required and when the trade-off between decreased computation time and increased memory usage is desired.

The top row of the table (Fig. \ref{fig:main}) mentions search-based methods, such as RRT \cite{lavalle1998rapidly}, PRM \cite{kavraki1996probabilistic}, and their numerous variants, have the goal of efficiently searching the free space in a planning environment to find a path that connects a starting point to a desired point. Most modern approaches can be assigned to this category, as they combine sampling search strategies and finite spatial representations with trajectory optimization for fast and reliable path planning \cite{sundaralingam2023curobo, cohn2023non, thomason2023motions} and others.

On the other hand, PDE based methods for path planning are a subcategory of potential field methods \cite{khatib1986real}, and these aim to compute a function in the domain that defines a gradient related policy, leading to reactive motions that are repelled from obstacles and attracted to a goal point or set. In general there are no closed form solutions for PDEs in arbitrarily domains. Koditschek and Rimon \cite{koditschek1990robot} describe multiple desired properties that potential fields should have for navigation, and refers functions fulfilling these conditions as navigation functions \cite{lavalle2006planning}. A multiple query method to solve a wide array of PDEs for motion planning is to use Physics Informed Neural Networks (PINN) \cite{ni2023progressive, ni2022ntfields, karniadakis2021physics, almqvist2021fundamentals}. The method has a heavy offline computational aspect, where the neural network encoding the solution of a PDE is solved iteratively through optimization in an offline training setting. The gradient of the neural networks encoding the solution is queried during the usage of the method to use the solution as a potential field for navigation.

The lower left quadrant in the table is the main focus of this paper. To the author's knowledge, no method has been introduced to solve PDEs for path planning as a single query or online approach, i.e., without solving the function in the entire domain, that is sampling based, exact in expectation, and probabilistically complete in a planning sense. In contrast to numerical methods that require an explicit representation of the domain, WoS relies on an implicit representation using a distance function (distance to the domain boundary). Walk on Spheres (WoS) is a method to compute the solution and solution-gradient of linear elliptic PDEs at a given point in the domain \cite{sawhney2020monte, muller1956some}. Although WoS itself is a PDE solver and not a path planning method, we consider it in this categorization as a single query, PDE based method, given that it is the enabler and main component of the approach. In addition, the algorithm is trivially parallelizable and can be used to solve screened Poisson equations, which we argue are highly useful as a parametric set defining potential fields for navigation.

\section{Screened Poisson Equation} \label{sec:screened_poisson}

The general screened Poisson equation with Dirichlet boundary condition can be expressed as
\begin{align}
    u - t\Delta u = f \quad & \mathrm{on} \quad \Omega \notag \\
    u = g \quad & \mathrm{on} \quad \partial\Omega \label{eq:screened_general}
\end{align}
with a parameter $t \in \mathbb{R}_{\geq 0}$, source function $f:\Omega \to \mathbb{R}$, and boundary function $g:\partial \Omega \to \mathbb{R}$, defined in the domain $\Omega \subseteq \mathbb{R}^n$ and its boundary $\partial \Omega$. The Poisson equation is the special case of the screened Poisson equation in the limit as $t\rightarrow\infty$.

An artificial potential field for navigation can be implicitly defined by the screened Poisson equation
\begin{align} \label{eq:screened_potential_field}
    u - t \Delta u = 0 \quad & \mathrm{in} \quad \Omega \notag \\
    u = 1 \quad & \mathrm{on} \quad \partial\Omega,
\end{align}
where the solution $u$ is positive and differentiable in the domain interior, and has a unique maximum at the boundary representing the goal set.

Using the transformation
\begin{align} \label{eq:transformation_varadhan}
    w(\mathbf{x}) = -\sqrt{t} \,\mathrm{ln}[u(\mathbf{x})]
\end{align}
the potential field $u(\mathbf{x})$ can be equivalently described as the solution $w(\mathbf{x})$ to the regularized Eikonal equation
\begin{align} \label{eq:regularized_eikonal}
    (1 - |\nabla w|^2) + \sqrt{t} \Delta w = 0 \quad & \mathrm{in} \quad \Omega \notag \\
    w = 0 \quad & \mathrm{on} \quad \partial \Omega,
\end{align}
\cite{belyaev2015variational, gurumoorthy2009schrodinger}.

From the transformation \eqref{eq:transformation_varadhan} it results that the gradients of the functions $u$ and $w$ are parallel everywhere in the domain. This implies that a gradient ascent policy on the solution to \eqref{eq:screened_potential_field} can be used to guide particles in the domain to the goal set.

We define \textit{screened harmonic paths} as the set of paths traversed through gradient ascent (or continuous gradient integration) by a particle in the domain, depending on the $t$-parameter
\begin{align}
    \nabla_\gamma \mathbf{x}_t = \frac{\nabla_\mathbf{x} u(\mathbf{x}_t(\gamma))}{|\nabla_\mathbf{x} u(\mathbf{x}_t(\gamma))|} \notag \\
    \mathbf{x}_t(0) = \mathbf{x}_{t,0}, \quad \mathbf{x}_t(1) = \mathbf{x}_{t,1} \notag \\
    0 \leq \gamma \leq 1, \quad \gamma \in \mathbb{R},
\end{align}
where $\mathbf{x}_t(\gamma)$ is the parametric equation of the path.

The resulting behavior as a function of the $t$ parameter can be derived from \eqref{eq:regularized_eikonal}. The Poisson equation is recovered in the limit as $t\rightarrow\infty$, and leads to smooth and boundary repulsive or Voronoi biased paths. In the limit as $t\rightarrow0$ the Eikonal equation is recovered, which leads to the shortest path, which is generally non-smooth and can traverse tangentially to domain boundaries.

Alternatively, the zero level-set of the distance function can be encoded in the source function,
\begin{align}
    u_t - t \Delta u_t = u_0 \quad & \mathrm{in} \quad \Omega \notag \\
    u_t = 0 \quad & \mathrm{on} \quad \partial\Omega,
\end{align}
where $u_t$ represents the heat in the domain interior after a diffusion time $t$, and an initial heat distribution $u_0$ according to Crane et al. \cite{crane2013geodesics}.

To match the notation used by Sawhney and Crane \cite{sawhney2020monte}, we set the parameter $c=1/t$ and regard the PDE with zero Dirichlet boundary conditions
\begin{align}
    \Delta u - c u = f \quad & \mathrm{in} \quad \Omega \notag \\
    u = 0 \quad & \mathrm{on} \quad \partial \Omega \label{eq:screened_poisson_wos}
\end{align}
as the assumed form screened Poisson equation in the rest of the paper.

\section{Walk on Spheres} \label{sec:wos}

In this section, we introduce the method \textit{Walk on Spheres} and the specific application to solve the screened Poisson equation. For a more detailed and complete treatment of the method, refer to \cite{muller1956some, sawhney2020monte}. 

Walk on Spheres (WoS) is a grid-free algorithm used to solve boundary value problems (BVP) as linear elliptic PDEs. WoS is a Monte Carlo method, so it computes exactly, in expectation, the solution of the BVP at a given point in the domain. As a Monte Carlo method, WoS relies on the Feynman-Kac formula, which relates PDEs to stochastic processes. 

For the Laplace problem, Kakutani's Principle \cite{kakutani1944143} states: 
\begin{quotation}
The solution value $u(\mathbf{x})$ at any point $\mathbf{x} \in \Omega$ is equal to the expected value $\mathbb{E}[g(y)]$, where $y \in \partial \Omega$ is the first boundary point reached by a random walk starting at $\mathbf{x}$ (more formally, a \textit{Wiener} Process) \cite{sawhney2020monte}
\end{quotation}

\begin{figure}
    \centering
    \includegraphics[width=0.9\linewidth]{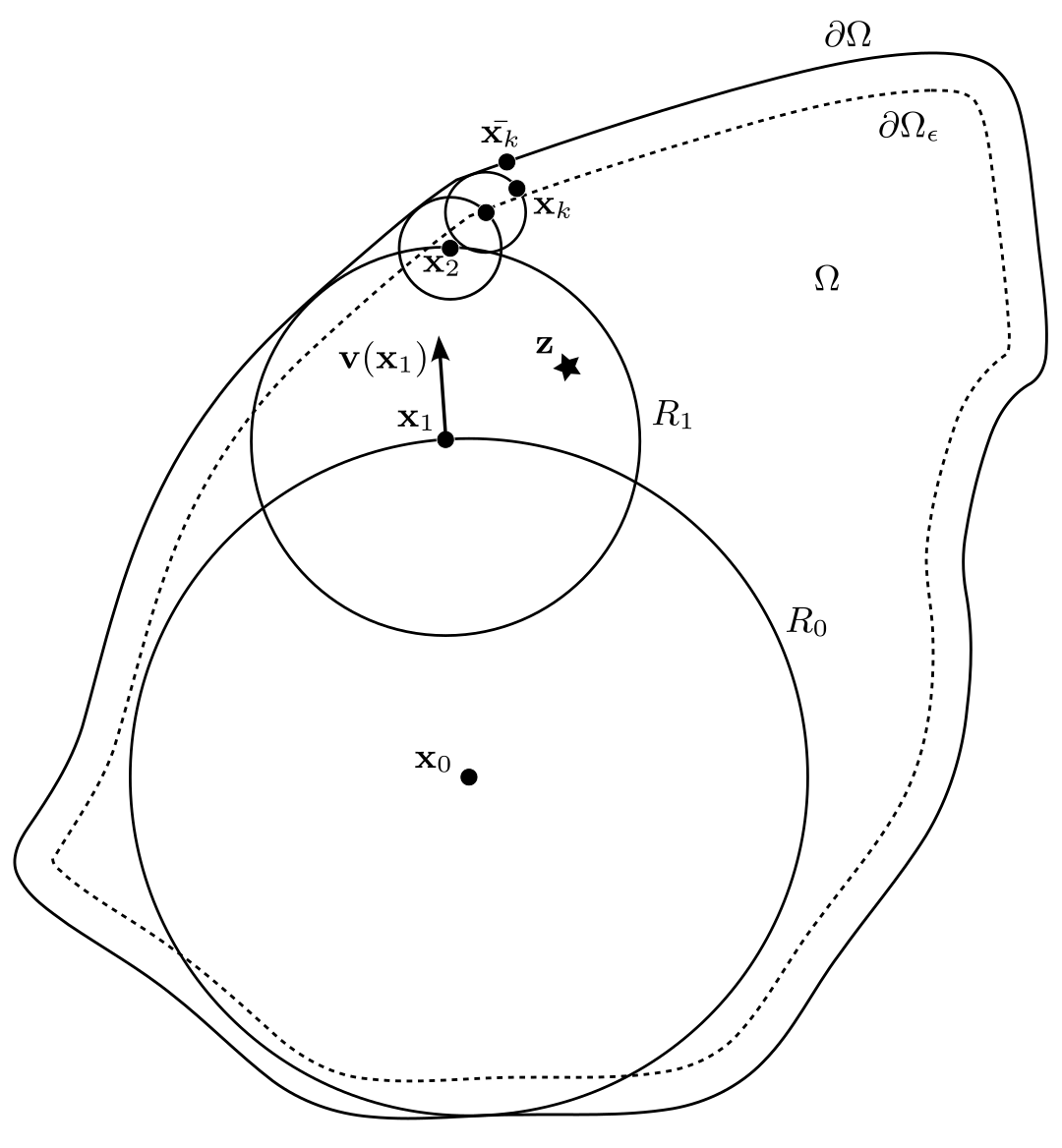}
    \caption{Walk on Spheres, algorithm visualization. The random walker starting at $\mathbf{x}_i$ is simulated by recursively jumping to a random uniformly sampled point $\mathbf{x}_1$ on the largest empty sphere (radius $R_i$) centered at $\mathbf{x}_i$. The boundary reaching point is approximated by $\bar{\mathbf{x}_k}$, which is the closest point on the boundary $\partial \Omega$ to the first sampled point ($\mathbf{x}_k$) at a distance lower then $\epsilon$ from the domain boundary. The unit vector $\mathbf{v}$ is used to estimate the gradient, and the source point $\mathbf{z}$ appears in the consideration of source function contributions, both for the value and gradient expressions.}
    \label{fig:wos_figure}
\end{figure}

Instead of explicitly simulating random walkers to compute this expectation, WoS leverages the following idea: a random walker that starts at the center of a ball domain, is equally likely to leave the domain through all points of the spherical boundary, and perfectly simulates the random walker through a \textit{Spherical Process} -- recursive sampling on the largest empty sphere centered at the query point. The boundary value at the boundary reaching point of a simulated random walker serves as a one-point estimate of the expectation integral and is the WoS estimator for the Laplace problem. The Spherical Process to simulate random walks is visualized in Figure \ref{fig:wos_figure}.

\subsection{WoS for the Screened Poisson Equation}

The solution to the screened Poisson equation with Dirichlet boundary conditions \eqref{eq:screened_poisson_wos} can be formulated as a recursive integral equation that generalizes the mean value property for the screened Poisson equation, and can be computed with the WoS method as the mean of the one-point estimates $\widehat{u}$  
\begin{align}
    \widehat{u}(\mathbf{x}_k) := \begin{cases}
g(\Bar{\mathbf{x}}_k),  \quad  \mathbf{x}_k \! \in \! \partial \Omega_\epsilon \\
C(\mathbf{x}_k)\widehat{u}(\mathbf{x}_{k+1}) + |B(\mathbf{x}_k)| f(\mathbf{y}_k) G_c(\mathbf{x}_k, \mathbf{y}_k),
\end{cases}
\end{align}
where the function $B(\mathbf{x})$ is the largest empty ball centered at $\mathbf{x}$ and its volume is represented as $|B(\mathbf{x})|$, the function $G_c(\mathbf{x}, \mathbf{y})$ denotes the Yukawa potential on the ball $B(\mathbf{x})$. The normalization constant $C(\mathbf{x})$ depends on the domain's spatial dimension and the radius of the largest empty ball centered at $\mathbf{x}$ \cite{sawhney2020monte}.

The gradient of the solution can be expressed by the recursive integral equation
\begin{align}
    \nabla u(\mathbf{x}) = \frac{1}{|B(\mathbf{x})|}\int_{\partial B(\mathbf{x})} u(\mathbf{y})\mathbf{v}(\mathbf{y})\mathbf{dy} + \notag \\ 
    \int_{B(\mathbf{x})}f(\mathbf{y}) \nabla G_c(\mathbf{x}, \mathbf{y}) \mathbf{dy},
\end{align}
where the vector $\mathbf{v} (\mathbf{x})$ is of unit length and normal to the sphere $\partial B(\mathbf{x})$, and pointing radially outwards (see Fig. \ref{fig:wos_figure}). The gradient of the solution can be computed through mean value principle
and its one-point WoS estimator 
\begin{align}
    \widehat{\nabla u_0}(\mathbf{x}_0) := \frac{n}{R} \widehat{u}_0(\mathbf{x}_1)\mathbf{v}(\mathbf{x}_1) + c_{\mathrm{imp}}\nabla G_c(\mathbf{x}_0, \mathbf{z}) \,\mathds{1}\{d_z<R_0\} \notag \\ \mathbf{x}_1 \sim \mathcal{U}(\partial B (\mathbf{x}_0)),
\end{align}
where $n/R$ is the surface area to volume ratio of an $n$-dimensional ball, and $c_\mathrm{imp} \in \mathbb{R}_{>0}$ is a positive importance sampling parameter \cite{sawhney2020monte}. Note that $\nabla G_c(\mathbf{x}_0, \mathbf{z})$ is zero when the distance between a point $\mathbf{x}$ and the source point $\mathbf{z}$ is larger than $R$, the distance from $\mathbf{x}$ to the boundary $\partial \Omega$. In the example walk visualized in Figure \ref{fig:wos_figure} the case $\mathds{1}\{d_z<R_0\}$ arises for the step centered at $\mathbf{x}_1$.

\subsection{Convergence and Complexity} \label{sec:wos_convergence}
A desirable quality of the WoS algorithm is its sound scaling in computation and memory effort as the spatial dimensionality increases, in contrast to its discretization-based counterparts. 
The order of convergence of the variance is independent of the spatial dimension and is expected to converge in the order of $O(1/n)$, and the standard deviation in the order of $O(1/\sqrt{n})$, where $n$ is the number of one-point estimates, i.e., the number of walks \cite{sawhney2020monte}. Further, the increase in computation effort as a function of the dimensionality is approximately linear in the dimensionality, hence, the utility of the WoS method increases with the dimensionality of the problem \cite{muller1956some}.

\subsection{WoS Applications in Path Planning} \label{sec:applications_wos}

As an application in Path Planning, we propose to use the WoS method to compute artificial potential fields for navigation that are implicitly defined by the screened Poisson equation introduced in Section \ref{sec:screened_poisson}. For this purpose, we make use of the following reformulation and alternative derivation.

The zero level-set of the distance function can be encoded in the source function,
\begin{align}
    u_t - t \Delta u_t = u_0 \quad & \mathrm{in} \quad \Omega \notag \\
    u_t = 0 \quad & \mathrm{on} \quad \partial\Omega,
\end{align}
where $u_t$ represents the heat in the domain interior after a diffusion time $t$, and an initial heat distribution $u_0$. The solution $u_t$ is equal to the solution of \eqref{eq:screened_potential_field} up to a multiplicative constant for a point source \cite{crane2013geodesics}.

To further match the notation of \eqref{eq:screened_poisson_wos} used in \cite{sawhney2020monte}, we set the parameter $c=1/t$ and regard the PDE with zero Dirichlet boundary conditions in form \eqref{eq:screened_poisson_wos} as the reference form screened Poisson equation in the rest of the paper.

The following properties motivate the usage of WoS in path-planning applications: 

\subsubsection{Trivially parallelizable} The expected computation error decreases with the number of walks, implying that the accuracy can be improved by a parallel increase in computation effort without increasing the computation time.

\subsubsection{PDE-Based} Harmonic paths and approximate shortest paths, with goal sets (more than one goal point), can be formulated as PDEs that are solvable by the WoS algorithm.

\subsubsection{Algorithmically independent of the spatial dimension} Path planning for robotic manipulators are usually high dimensional problems ($\sim 7$) and we have seen the success of sampling-based planners that share this quality.

A robotic application in the realm of holonomic path planning that is enabled by WoS, is the online computation of the gradient ascent direction on a navigation function landscape in higher dimensions ($>3$). We focus on this application to empirically evaluate the WoS method and its mentioned properties in a path planning environment.

\section{C-Space Distance-to-Boundary Function} \label{sec:lipschitz}

The WoS method relies on a query of the distance-to-boundary function $d(\mathbf{x}, \partial \Omega)$. However, when dealing with configuration spaces, a closed form solution or explicit representation is rarely available. Instead, implicit representations using a function $h: \mathbb{R}^n \to \mathbb{R}$ define the domain interior and its boundary
\begin{align}
    \Omega &= \{ \mathbf{x} \in \mathbb{R}^n \ | \ h(\mathbf{x}) \geq 0 \} \\
    \partial \Omega &= \{ \mathbf{x} \in \mathbb{R}^n \ | \ h(\mathbf{x}) = 0 \}.
\end{align}
Given that approximations of the distance-to-boundary function are valid as long as they underestimate the true distance, we approximate the distance-to-boundary based on the following observation: Let $h$ be \textit{Lipschitz continuous} in $\Omega$, then there exists a real constant $K\geq0$ such that, for all $\mathbf{x}_1, \mathbf{x}_2 \in \Omega$,
\begin{align}
    \lVert h(\mathbf{x}_2) - h(\mathbf{x}_1) \rVert \leq K \lVert \mathbf{x}_2 - \mathbf{x}_1 \rVert.
\end{align}
With the Euclidean distance in both input and output spaces of $h$, the tightest Lipschitz constant $K_\Omega$ for $h$ is equivalent to the largest norm of its gradient
\begin{align}
    K_\Omega = \mathrm{sup} \left\{ \left\| \frac{\partial h}{\partial \mathbf{x}}(\mathbf{x}) \right\| \ \Big| \  \mathbf{x} \in \Omega \right\}.
\end{align}

This construction guarantees that a ball
\begin{align}
    B\left(\mathbf{x}, \frac{h(\mathbf{x})}{K_\Omega} \right) \subseteq \Omega, \forall \mathbf{x} \in \Omega \label{eq:safe_ball},
\end{align}
i.e., lies entirely in the domain $\Omega$. Its radius $r=h(\mathbf{x})/K_\Omega$ can be used to underestimate the distance-to-boundary between the ball center point $\mathbf{x} \in \Omega$ and the domain boundary $\partial\Omega$.

\subsection{Distance-to-Boundary for Manipulators} \label{sec:lipschitz_manip}

We construct a simple and efficient underestimate for kinematic chains, based on the Lipschitz constant of the forward kinematic map (FKM), assuming static obstacles and using the task space distance function $d_\tau$ to implicitly define the domain. Note that the task space distance between the manipulator and static obstacles cannot vary faster than the fastest point on the manipulator's surface, which allows us to construct the following approximation.

The forward kinematic map of a robotic manipulator -- consisting of sequential rigid links and revolute joints -- is smooth everywhere, which implies its Lipschitz continuity. Because the FKM is a multivariate function, the Lipschitz constant $K$ is the largest singular value of the FKM-Jacobian in the domain. The FKM and therefore its Jacobian are determined by the manipulator's geometry, meaning that the Lipschitz constant can be obtained directly from the manipulator's DH parameters or similar representations. In the case of a planar manipulator, a Lipschitz constant is given by 
\begin{align}
    K = \sqrt{\sum_{n=1}^N \left(\sum_{i=1}^n l_i\right)^2},
\end{align}
where $N$ is the number of joints (and links) and $l_i$ is the length of each link \cite{lavalle2006planning}. The same formula can be used for general manipulators with revolute joints nevertheless potentially leading to a less tight estimate.

\begin{figure}[h]
    \centering
    \includegraphics[width=\linewidth]{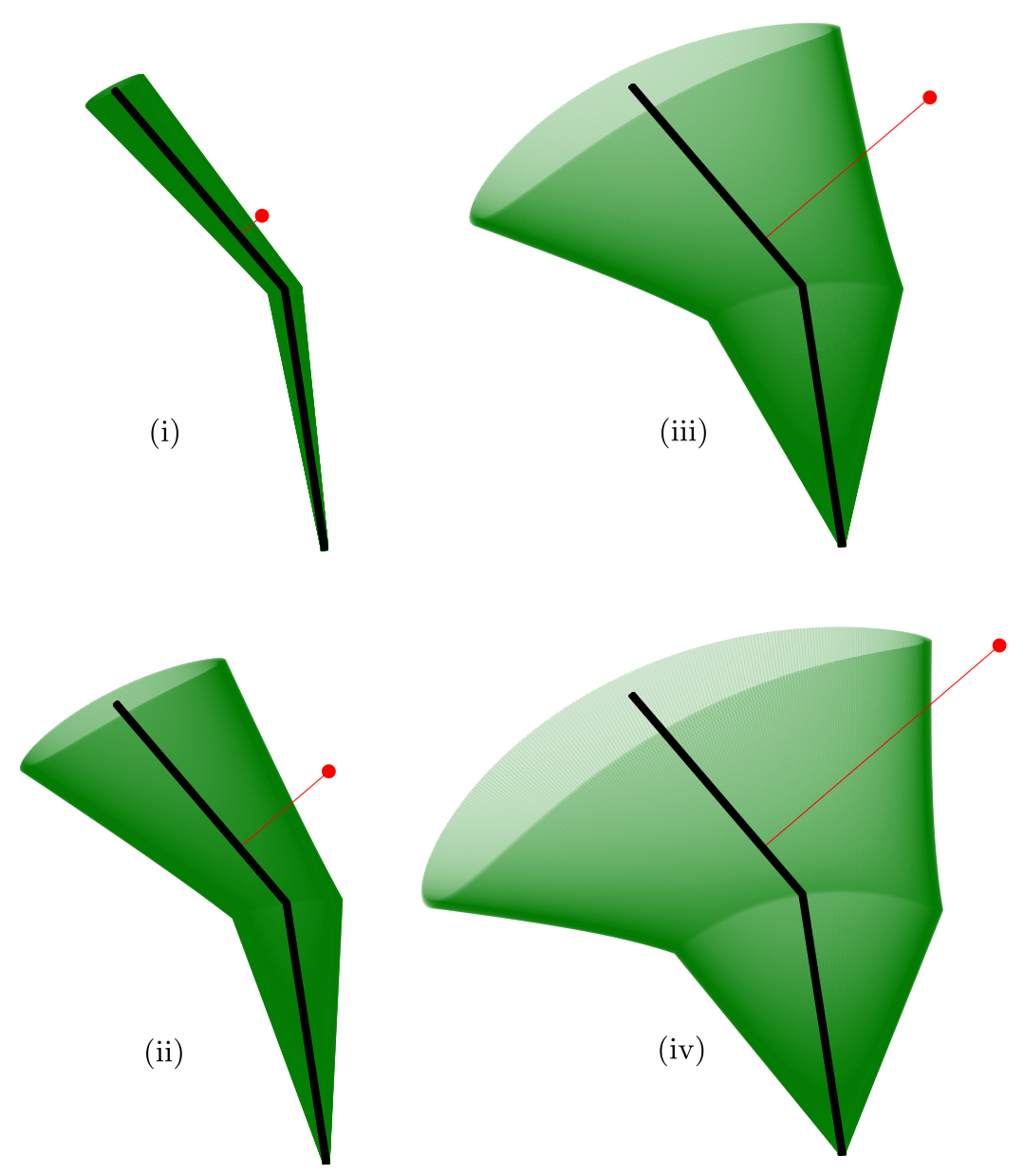}
    \caption{The task space distance $d_\tau(\mathbf{x})$ is the length of the red line between the RR planar manipulator in configuration $\mathbf{x}$ (black) and a point obstacle (red), using four obstacle locations with increasing distance. In each quadrant, the boundary circle of the approximated safe disk of radius ($K^{-1}d_\tau(\mathbf{x})$) is visualized through the task space representation of $500$ equidistant points (light green) on the boundary circle.}
    \label{fig:lipschitz}
\end{figure}

Simply, and underestimating by construction, we approximate the joint space distance-to-boundary function $d_\mathcal{Q}$ by
\begin{align}
    \tilde{d}_\mathcal{Q} = \frac{1}{K} d_\tau \leq d_\mathcal{Q}.
\end{align}
The resulting approximation of safe balls with this approach is visualized using a two-dimensional planar robot in Fig. \ref{fig:lipschitz}. In Section \ref{sec:validation_lipschitz} we leverage this Lipschitz-based approximation in a validating example.

\section{Empirical Evaluation} \label{sec:empirical_eval}

In this section, we evaluate the performance and the behavior of the WoS method, when solving the screened-Poisson equation at a given point in an environment with a non-convex obstacle. In this evaluation, we assess the following qualities of the algorithm: global, parameterized, parallel, online computation, and sound scaling into higher dimensions.

\begin{figure}
    \centering
    \includegraphics[width=0.8\linewidth]{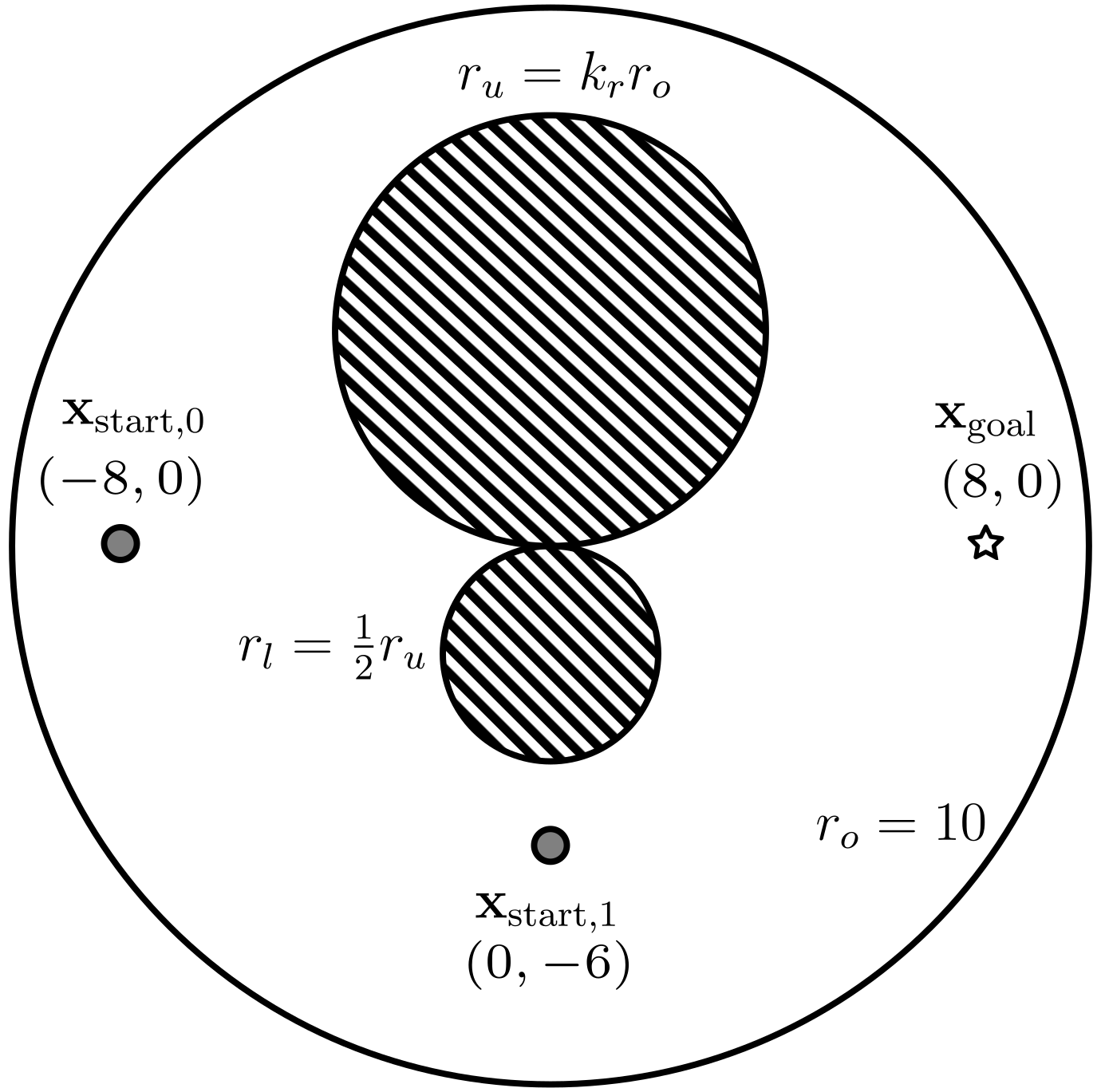}
    \caption{Notation used to describe the WoS evaluation environment.}
    \label{fig:notation_eval}
\end{figure}

The environment structure and notation are visualized in Figure \ref{fig:notation_eval}. The environment is determined by three disks in 2-dimensional Euclidean space -- an outer boundary of radius $r_o$ centered at the origin, an upper obstacle of radius $r_u$, and a lower of radius $r_l$. The boundary of the obstacles coincide only at the origin, forming a non-convex obstacle. The outer radius is fixed; the obstacle radii are determined by
\begin{align}
    r_o=10, \quad r_u=k_rr_o, \quad r_l=\frac{1}{2}r_u,
\end{align}
where the radius scaling parameter $k_r$ is used to vary the visibility (proxy for ease of planning, or inverse clutteredness) of the environment. The source function is defined as a unit impulse centered at the goal point $\mathbf{x}_\mathrm{goal}$. To evaluate the method in higher dimensions, the boundary and obstacles are extruded along the third dimension, and further unbounded dimensions are ``appended'' to the free space.

Given the simple and scalable geometry of the environment, the distance-to-boundary function is computed in the first two dimensions as
\begin{align}
    d(\mathbf{x}, \partial \Omega) = \mathrm{inf} \left\{ 
        r_o - \lVert \mathbf{x} \rVert_2,\ \lVert \mathbf{x} - \mathbf{c}_{i=u,l} \rVert_2 - r_{i=u,l}\right\},
\end{align}
where $\mathbf{c}_i$ represents the center point of the $i$th obstacle.

\begin{figure}
    \centering
    \includegraphics[width=\linewidth]{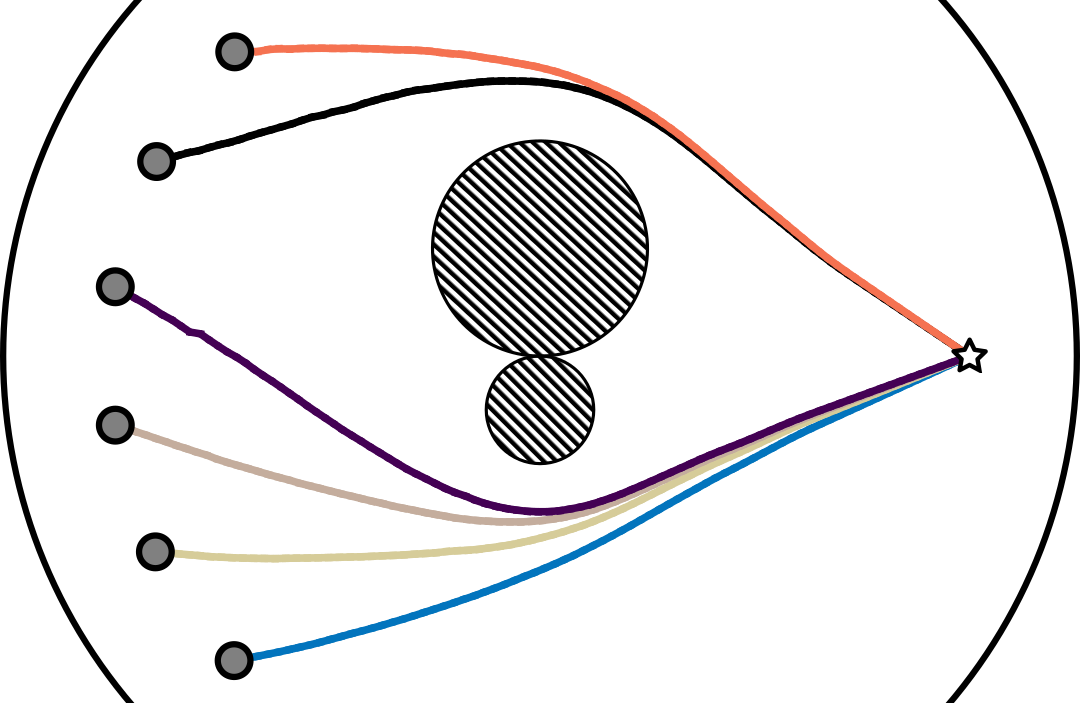}
    \caption{Demonstrating the landscape through multiple start points. Screening coefficient fixed to $c_{\mathrm{screen}} = 1$ for comparability. Radius scaling parameter $k_r=0.2$.} 
    \label{fig:multistart}
\end{figure}

Example paths from multiple starting points to the goal point in the evaluation environment are visualized in Figure \ref{fig:multistart} which shows the global quality of the algorithm and the smoothness of the computed screened-harmonic paths. At first, we evaluate the algorithm in the planar environment, and then extend the domain to higher dimensions to study its scaling characteristic in the final evaluation experiment.

\subsection{Screening Parameterization}
\begin{figure}
    \centering
    \includegraphics[width=\linewidth]{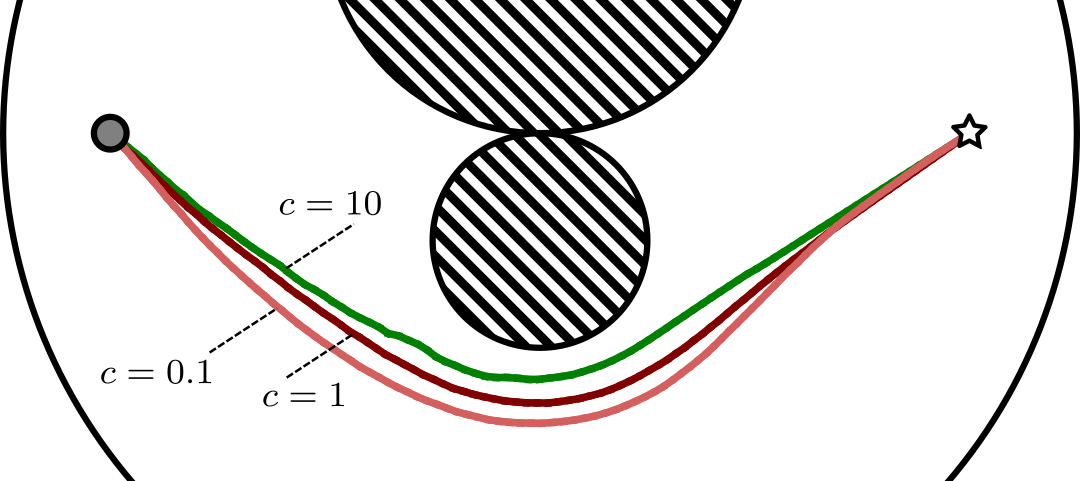}
    \caption{Screened Poisson Navigation. Screening coefficient $c_{\mathrm{screen}} = [10,\, 1,\, 0.1]$. This figure shows empirically, how the screening parameter $c_{\mathrm{screen}}$ affects the PDE solution: interpolate between shortest and harmonic paths. Obstacle size parameter $kv=0.4$.} 
    \label{fig:c_sweep}
\end{figure}

The screening coefficient $c$ of the screened Poisson equation is used as a parameter in path planning applications, that controls the behavior of the resulting potential field and therefore the integrated path. As presented in section \ref{sec:screened_poisson}, we expect the length of the path to decrease as the screening coefficient increases, given fixed start and goal points. To visualize this effect, we compute paths through gradient ascent integration using three screening coefficients $c_\mathrm{screen}=[10,\,1,\,0.1]$ while fixing all other parameters. The resulting paths are shown in Figure \ref{fig:c_sweep}.

As expected, the path length monotonically decreases with an increase of the screening coefficient. Further, the smoothness of the path decreases as well, which is likely due to the reduced distance to the obstacle that is also observed in the following visibility analysis -- Sec. \ref{sec:eval_visibility}. 

\subsection{Visibility and Number of Walks} \label{sec:eval_visibility}

\begin{figure}
    \centering
    \includegraphics[width=\linewidth]{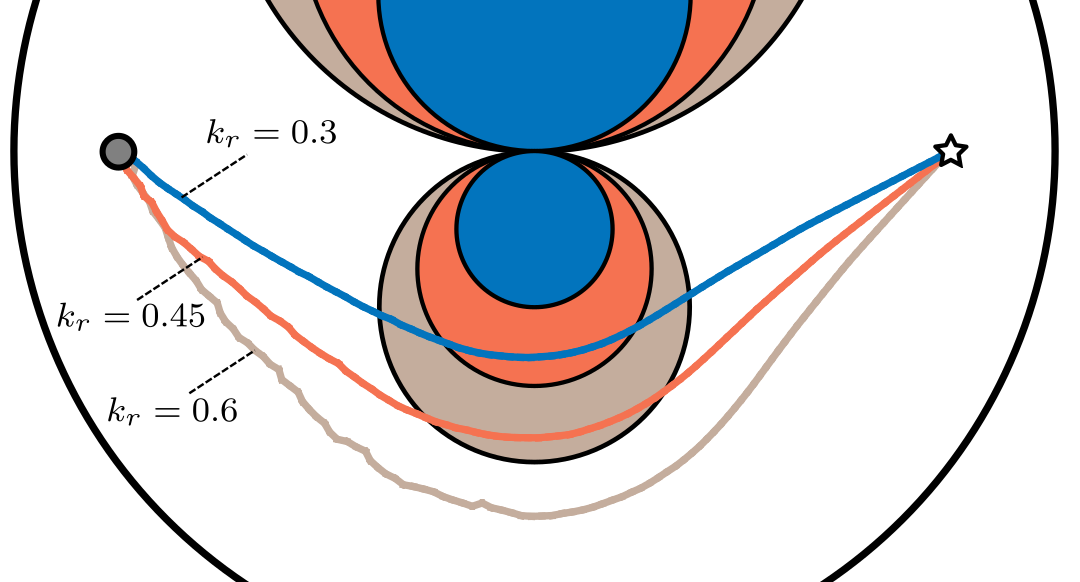}
    \caption{Effect of the environment visibility on the integrated path and estimate variance. Evaluated obstacle radius scales $k_r = [0.3,\, 0.45,\, 0.6]$. Screening coefficient fixed to $c_{\mathrm{screen}} = 1$ for comparability. Number of walks $n=10^5$. The variance of the step-wise gradient estimates increases with the "clutteredness" of the environment.} 
    \label{fig:kr_sweep}
\end{figure}

We observe the effect of the environment visibility on the integrated path. As the visibility decreases, i.e., the clutteredness increases, we expect less smooth integrated paths, arising from the reduced expected number of walks leading to a source contribution. To observe this effect we vary the radius scaling parameter $k_r$ that is inversely related to the visibility. The resulting environment structures and paths for the settings $k_r = [0.3,\, 0.45\,\ 0.6]$ are overlayed using respectively blue, orange, and brown colors in Figure \ref{fig:kr_sweep}.

Similar to the effect observed in the screening coefficient analysis, the smoothness of the path decreases with the visibility of the environment, i.e., with an increasing $k_r$ parameter. This result matches the theory and intuitive expectation, that a higher number of walks is necessary in a more cluttered environment to reach the same variance in the estimate.

\begin{figure}
    \centering
    \includegraphics[width=\linewidth]{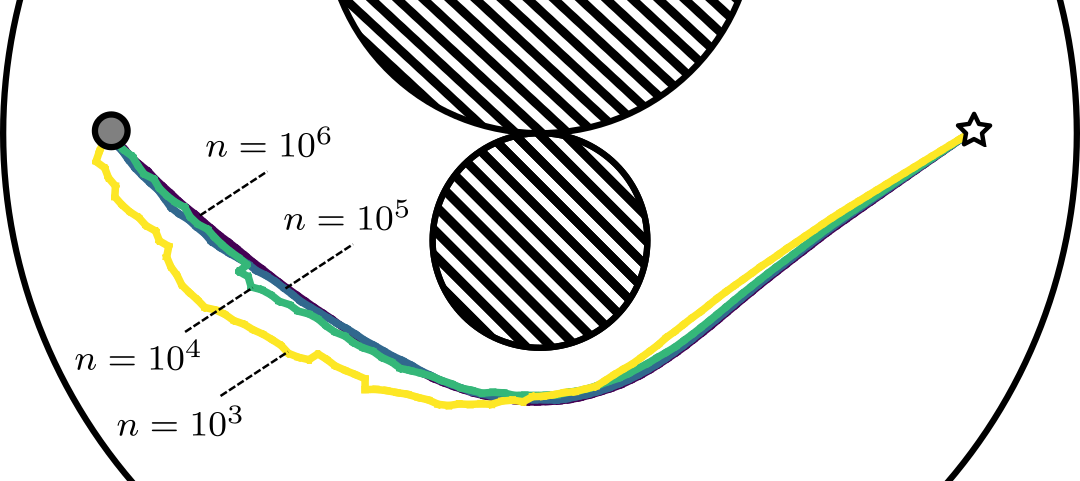}
    \caption{Effect of number of walks $n$ on the resulting integrated path. Evaluated number of walks $n = [10^3,\, 10^4,\, 10^5,\, 10^6]$. Screening coefficient fixed to $c_{\mathrm{screen}} = 1$ for comparability. Obstacle size parameter $kv=0.4$. The variance of the step-wise gradient estimates decreases as the number of walks increases.} 
    \label{fig:nwalks_sweep}
\end{figure}

To visualize the relationship between the number of walks and the estimate variance we evaluate the method with $n_\mathrm{walks}$ between $10^3$ and $10^6$, while fixing the further parameters. The resulting paths are visualized in Figure \ref{fig:nwalks_sweep}.

From Figure \ref{fig:nwalks_sweep} the trend of decreasing variance with an increase in $n_\mathrm{walks}$ is clear, leading to a convergence of the paths to the exact solution. An additional noticeable trend in the visualized paths is that the path smoothness is not constant throughout the path, but instead it increases as the proximity to the goal decreases. This behavior again supports the observed relationship between the environment visibility and the estimate variance with a fixed number of walks.

\subsection{Parallelization}

\begin{figure}
    \centering
    \includegraphics[width=\linewidth]{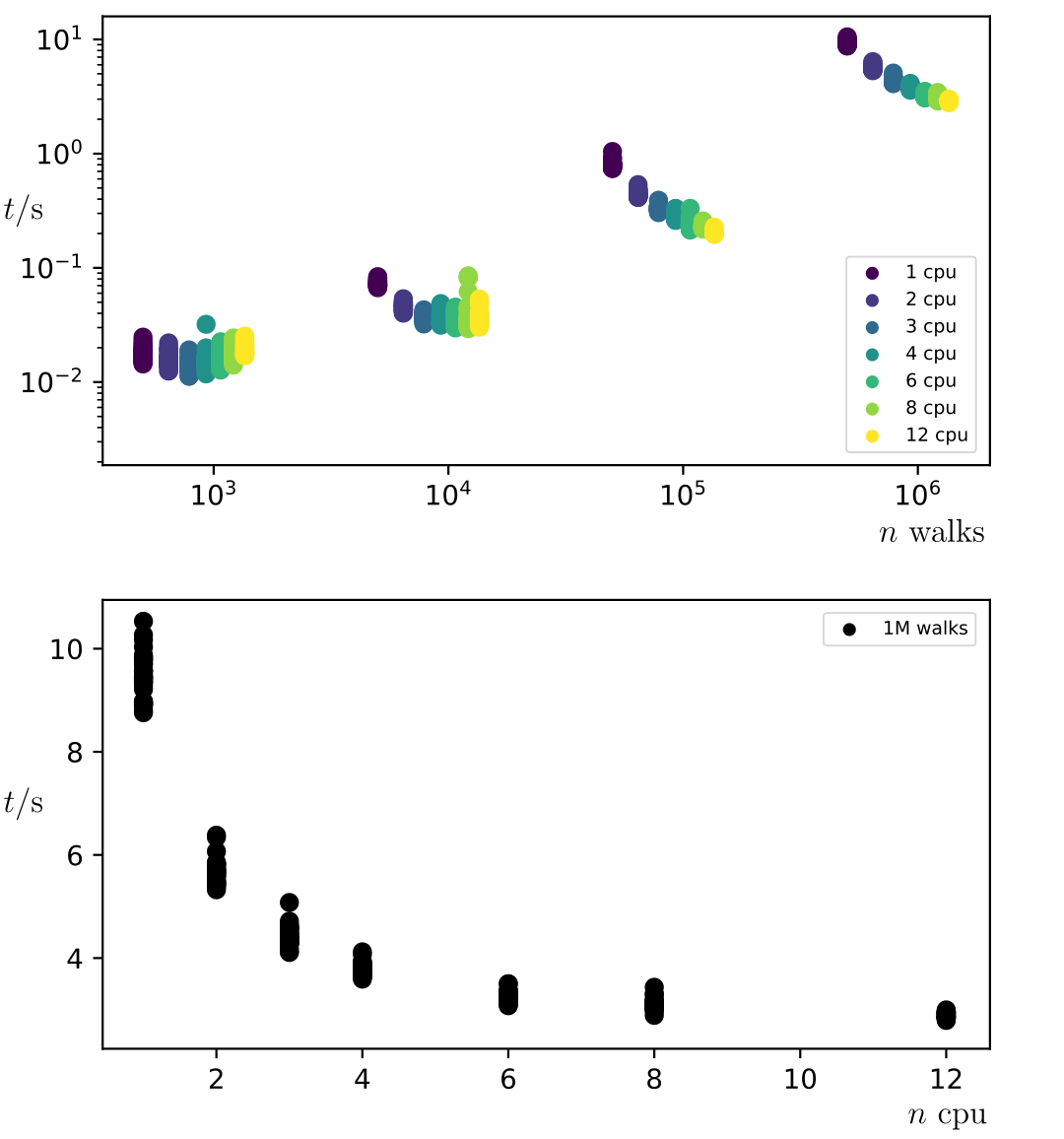}
    \caption{Empirical distribution of the computation time per time step, as a function of the number of CPU cores $n\,\mathrm{cpu}$ used in multiprocessing, and the number of walks per time step. Implemented with Python \textit{numpy} for vectorization, using the \textit{multiprocessing} module. Fixed parameters: 2-dimensional, $k_r=0.3$, $\epsilon=0.01$, $c_\mathrm{screen}=1$, collected $32$ data points per setting, point located at $(-8, 0)$ as in previous single-particle paths, processor ``Intel® Core™ i7-8750H CPU @ 2.20GHz × 12''}. 
    \label{fig:ncpus_sweep}
\end{figure}

The WoS method is trivially parallelizable, given that each walk can be traversed independently of the others, and the estimate is computed as an expectation over the walk endpoints and source contributions. This fact, combined with the evaluation in Figure \ref{fig:nwalks_sweep}, means that a higher expected estimate accuracy can be achieved without an increase in computation time by extending the parallel computation capacity. In a \textit{numpy} implementation, we compare the computation time depending on the number of CPU cores active in parallel ($1$-$12$) through the \textit{multiprocessing} package. The resulting computation times are visualized in Figure \ref{fig:ncpus_sweep}.

The trivial parallelization and the accompanying reduction in computation time are most noticeable with a high number of walks and low CPU cores. Although ideally, we expect a $\mathcal{O}(1/n_\mathrm{CPU})$ characteristic, the overhead in multiprocessing and the non-optimized Python implementation lead to a slowing reduction in computation time as the number of parallel processes increases.

\subsection{Dimensionality Scaling}

\begin{figure}
    \centering
    \includegraphics[width=\linewidth]{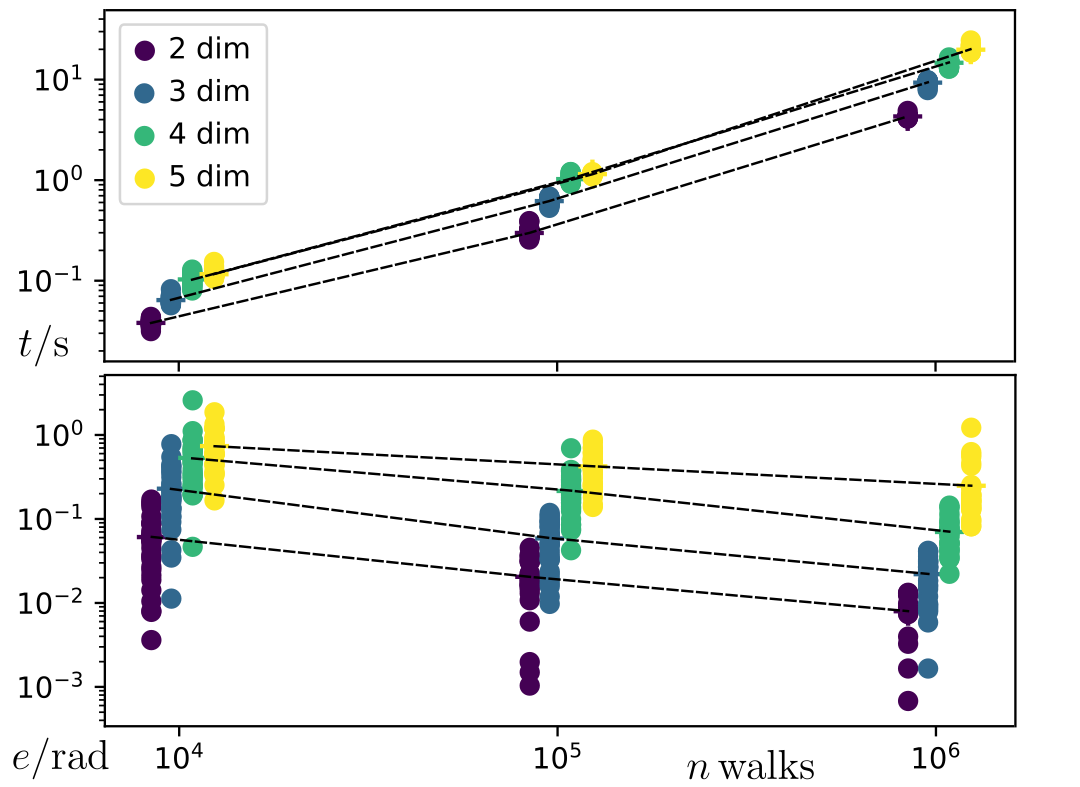}
    \caption{Dimensionality study with logarithmic y- and x-axes. Top: computation time (at D10 CPU cores). Bottom: error as angle error between estimated normalized gradient and ground truth, computed as the expectation at 100M walks for each dimension. The obstacles maintain their geometry and the dimensionality of the free space is increased and unbounded in each additional dimension. The dashed lines are plotted through the mean value at each dimension and number of walks. 
    The expected $O(1/\sqrt{n_\mathrm{walks}})$ convergence behavior and the dimensionality independence can be observed in the plot.} 
    \label{fig:dim_sweep}
\end{figure}

As mentioned in the beginning of this section, the environment dimensionality can be extended, while preserving the distance function and some intuition regarding its structure. In this evaluation, we measure the error and computation time at multiple settings, determined by the number of walks and the environment dimensionality. Given that for path planning applications we are interested in the gradient direction, we define the error $e$ as the angle between the estimated gradient and the exact direction, i.e.,
\begin{align}
    e &= \mathrm{arccos}(\frac{\hat{\mathbf{g}}^T \mathbf{g}}{\lVert \hat{\mathbf{g}} \rVert_2 \lVert \mathbf{g} \rVert_2 }) \\
    \mathbf{g} &= \nabla_\mathbf{x} u
\end{align}
where $\mathbf{g}$ is the solution's gradient. The point in the domain is fixed at $\mathbf{x}_{\mathrm{start},1}$ (see Fig. \ref{fig:notation_eval}). The results are shown in Figure \ref{fig:dim_sweep}.

The upper plot in Figure \ref{fig:dim_sweep} shows the relationship between the number of walks and computation times in spatial dimensions in the $2$-$5$ range. The characteristic of the computation time can be observed as $O(n_\mathrm{walks})$ and a scaling factor that correlates with the spatial dimension, shown by the linear relationship in the logarithmic plot. The lower plot in Figure \ref{fig:dim_sweep} presents the error as a function of the number of walks and the spatial dimension. The error is observed to decrease according to $O(1/\sqrt{n_\mathrm{walks}})$ and, again, a scaling factor that correlates with the spatial dimension.

These results match the expected theoretical characteristics presented in Section \ref{sec:wos_convergence} and motivate the further development of specialized path planning methods based on WoS.

\section{Validation Experiment on the RR Platform} \label{sec:validation_rr}

\begin{figure}
    \centering
    \includegraphics[width=\linewidth]{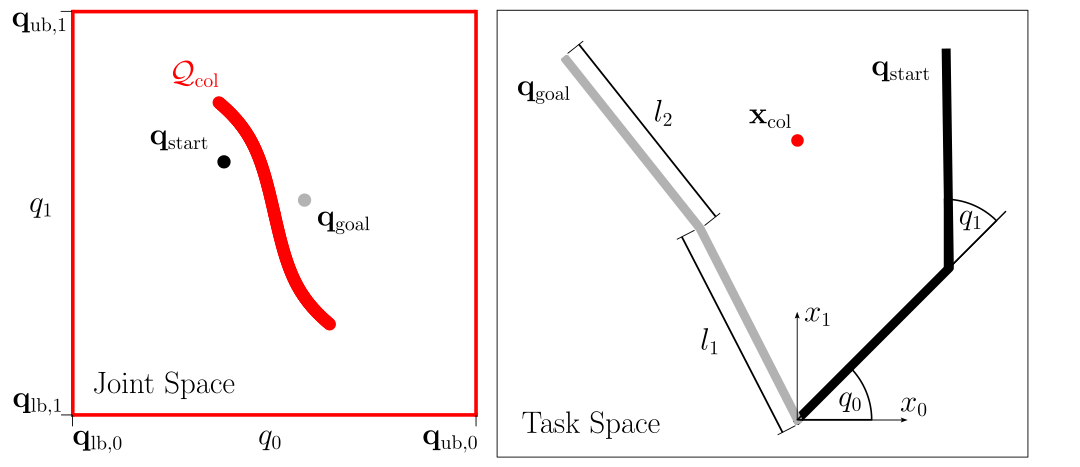}
    \caption{RR Environment: task structure and notation.}
    \label{fig:RR_notation}
\end{figure}

As a validation experiment on a simulated robotic platform, we use the WoS algorithm as an online policy for the RR robot. The task is defined as follows: given an initial configuration $\mathbf{q}_\mathrm{start}$, a goal configuration $\mathbf{q}_\mathrm{goal}$, and a point-obstacle in task-space $\mathbf{x}_{\mathrm{col}}$, find the screened harmonic path $\mathbf{q}_u(\gamma)$, in the free configuration space $\mathcal{Q}_\mathrm{free}$
\begin{align}
    \Delta_q u(\mathbf{q}) - cu(\mathbf{q}) = f(\mathbf{q}) \quad \mathrm{on} \quad \mathcal{Q}_\mathrm{free} \notag \\
    \nabla_\gamma  \mathbf{q}_u(\gamma) = \mathrm{normalized}(\nabla_q u(\mathbf{q}_u(\gamma))^T) \notag \\
    \mathbf{q}_u(0) = \mathbf{q}_\mathrm{start}, \quad \mathbf{q}_u(1) = \mathbf{q}_\mathrm{goal} \\
    0 \leq \gamma \leq 1
\end{align}
considering the task-space obstacle and configuration space boundaries.
We define the source function $f$ as the unit impulse centered at the goal configuration $\mathbf{q}_\mathrm{goal}$. The task environment and the notation are visualized in Figure \ref{fig:RR_notation}. In the discrete implementation, we set an adaptive integration step size $s$ as the minimum of a nominal step size $s_{\mathrm{ub}}$ and half the distance to the boundary $d_\mathrm{bound}$. This adaptive step size ensures that points in the domain that follow the gradient policy, also remain in the domain interior, effectively rendering the free configuration space forward invariant.

The configuration space boundaries are defined by the union of (i) a box constraint, i.e., constant upper and lower bounds for each joint, and (ii) the configurations $\mathcal{Q}_\mathrm{col}:=\mathcal{Q}\setminus \mathcal{Q}_\mathrm{free}$ in collision with the task-space point obstacle, which is a curve (bounded one-dim subset) in the two-dimensional $\mathcal{Q}_\mathrm{free}$. Using the analytical IK solution of the RR robot we obtain a closed-form expression describing the configurations in collision.

In this validation experiment we compute the distance-to-boundary function with two different methods (i) using a point-based closed form IK solution of the RR manipulator, and (ii) using the Lipschitz-based approximation presented in Section \ref{sec:lipschitz}.

\subsection{IK-based Distance Function}

We represent the set $\mathcal{Q}_\mathrm{col}$ as $N_\mathrm{col}$ approximately equidistant points $\mathbf{p}_{\mathrm{col},i}$ that lead to the distance function approximation as
\begin{align}
    \hat{d}(\mathbf{q}, \mathcal{Q}_\mathrm{col}) = \mathrm{inf}\left\{\, || \mathbf{q} - \mathbf{p}_{\mathrm{col},i} ||_2 \ \middle| \  \mathbf{p}_{\mathrm{col},i} \in \mathcal{Q}_\mathrm{col} ,\, i \in \mathbb{N}_{\leq N_\mathrm{col}} \right\},
\end{align}
i.e., the shortest Euclidean distance between a given configuration and the set of points representing the collision-free configuration space.

The euclidean distance in configuration space between consecutive points $\mathbf{p}_{\mathrm{col},i}$ and $\mathbf{p}_{\mathrm{col},i+1}$ ranges from $0.017$ to $0.0235$. This range guides the choice of the $\epsilon$-parameter in the WoS algorithm. The environmental and algorithmic parameters used during the simulation are shown in Table \ref{tab:rr_environment}.

\begin{table}[]
    \centering
    \begin{tabular}{|c|c|}
        \hline
        Parameter & Value \\ \hline \hline
        \rule{0pt}{2ex} $\mathbf{q}_\mathrm{ub}$ & $[3\pi/2, \ \pi]$ \\
        $\mathbf{q}_\mathrm{start}$ & $[0.785, \ 0.800]$ \\
        $\mathbf{q}_\mathrm{goal}$ & $[2.042, \ 0.200]$ \\
        $\mathbf{x}_\mathrm{obs}$ & $[0.0, \ 1.3]$ \\
        $N_\mathrm{col}$ & $200$ \\ 
        $n_\mathrm{walks}$ & $10^5$ \\
        $s_\mathrm{ub}$ & $0.1$ \\
        $s$ & $\mathrm{min}(s_\mathrm{ub}, 0.5 \,d_\mathrm{bound})$ \\
        $c_{\mathrm{screen},a}$ & $0.0$ \\
        $c_{\mathrm{screen},b}$ & $5.0$ \\
        $\epsilon$ & $0.02$ \\ \hline
    \end{tabular}
    \caption{RR Environment and WoS Parameters}
    \label{tab:rr_environment}
\end{table}

\begin{figure}[h]
    \centering
    \includegraphics[width=\linewidth]{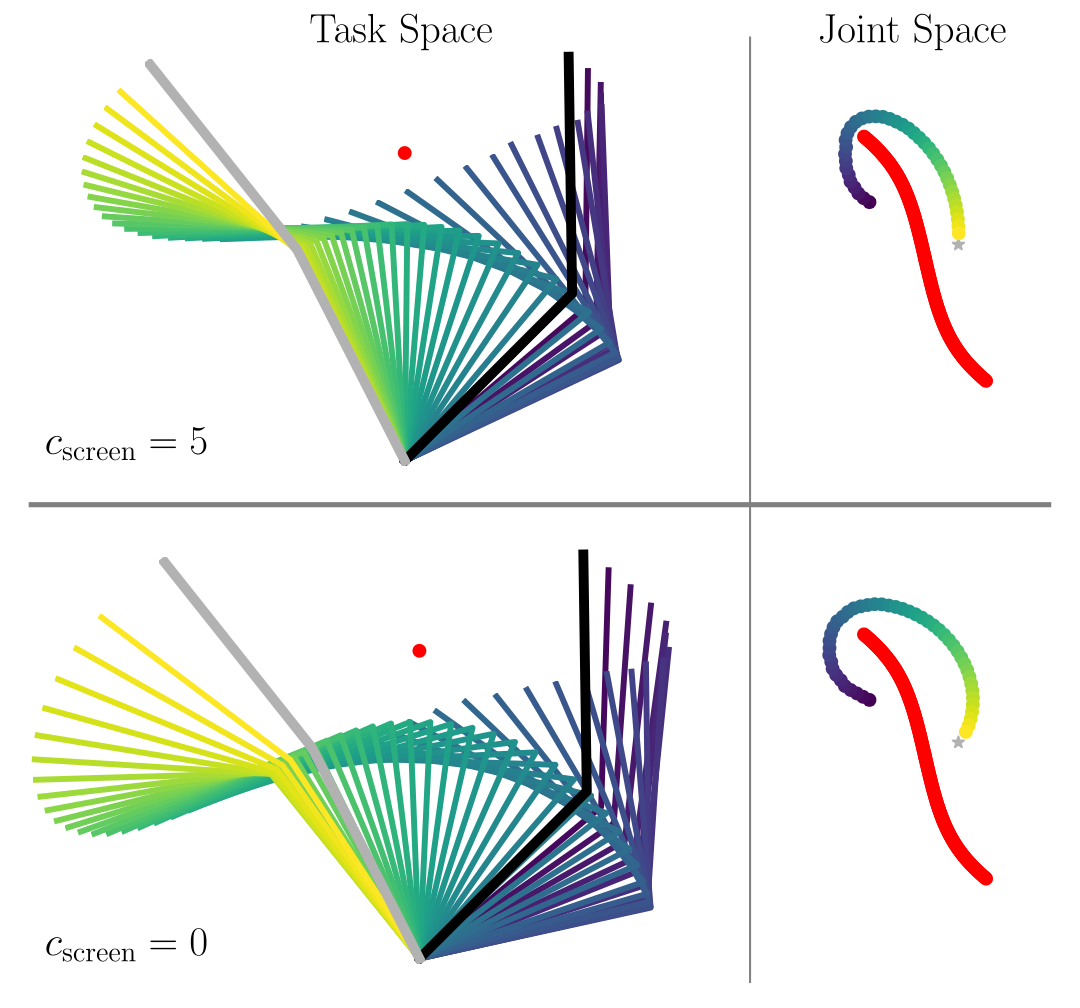}
    \caption{Validation experiment on the RR platform. Computed with $n_\mathrm{walks}=10^5$, step size $s=\mathrm{min}(0.1, 0.5d_\mathrm{bound})$ and $\epsilon=0.02$. The path lengths for $c=5$ and $c=0$ are correspondingly $3.8$ and $4.5$, quantifying the effect of the screening parameter on the solution.}
    \label{fig:rr_wos}
\end{figure}

The computed paths, both in joint- and task-space, are visualized in Figure \ref{fig:rr_wos}. The paths converge to the goal and follow the desired screened-harmonic behavior, which is smooth, local minimum-free, and obstacle-avoiding. We also see a similar behavior as in the empirical validation section, where an increase in the screening parameter leads to a shorter path ($3.8$ vs. $4.5$), and less accurate gradient estimates at a fixed number of walks.

\subsection{Lipschitz-based Distance Function} \label{sec:validation_lipschitz}

\begin{figure}
    \centering
    \includegraphics[width=\linewidth]{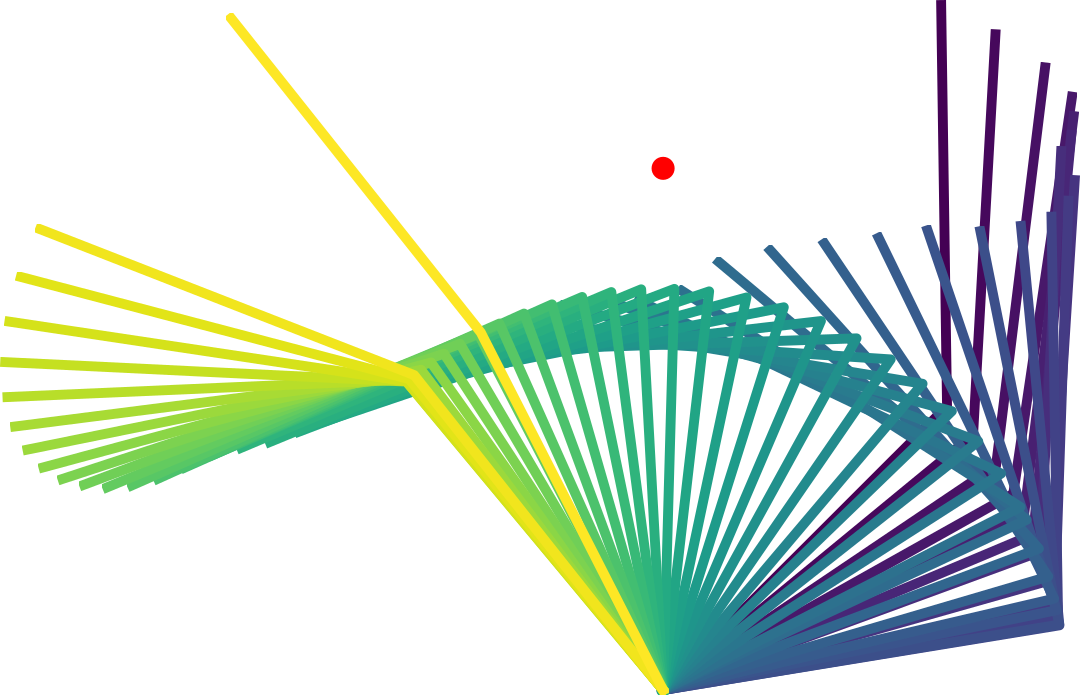}
    \caption{Path between configurations in the RR simulation environment, computed with the Lipschitz-based conservative distance approximation and the parameters $n_\mathrm{walks}=10^5$, $c=0$, step size $s=0.1$, and $\epsilon=0.02$. To visualize the distance function approximation, the final configuration lies within the Lipschitz ball centered at the penultimate configuration.}
    \label{fig:harmonic_lipschitz}
\end{figure}

To support the theoretical basis through an empirical test, we recompute the harmonic path in the RR environment as presented in section \ref{sec:validation_rr}, while using the Lipschitz-based conservative approximation of the joint space distance function described in Section \ref{sec:lipschitz_manip}. The resulting path is visualized in Fig. \ref{fig:harmonic_lipschitz} and coincides with the IK-based path computed in the previous section.

\section{Discussion and Limitations}

The WoS method simulates the random walks through Markovian spherical processes, as presented in Section \ref{sec:wos}, and does not make use of task-specific knowledge.
Similar to the development in RRT and related sampling-based planners, we are optimistic about methods that build upon WoS and allow us to incorporate task-specific knowledge through sampling biases (e.g. importance sampling) toward more efficient resource usage, overcoming initial limitations such as the increasing number of walks required in higher dimensions and the navigation through narrow or long passages.

In this work we focused on CPU-based parallelization; however, the use of GPUs presents a further possibility of scaling the results to real-world applications, and initial experiments of GPU-based WoS support this development \cite{mossberg2021gpu}.
Nevertheless, the trends resulting from the analyses and evaluations presented in this paper are specific to the WoS method for path planning and therefore independent of the underlying computing structure.

Multiple variants of the WoS method have been recently developed in the computer graphics community in an ongoing and growing effort to adapt the method to solve a wider range of PDEs. A few that are promising and useful in path planning applications are: a method that considers varying spatial coefficients \cite{sawhney2022grid}, Walk on Stars that enables Neumann boundary conditions \cite{sawhney2023walk}, and a bidirectional method \cite{qi2022bidirectional}.

The Lipschitz-based approximation of the distance to the domain boundary becomes more conservative as the dimensionality of the manipulator increases, which is a natural result given that it is a global property. Although this is not a theoretical limitation of the approach, practically it can lead to small steps and slow walks. Locally restricted Lipschitz constants could be used to overcome this potential limitation.

In this paper, we focus on holonomic systems and first-order control. This invites the challenge to develop highly parallelizable planning algorithms for kinodynamic and higher-order control systems.  

\section{Conclusion} 
\label{sec:conclusion}

In this paper we introduce a sampling-based method to compute artificial potential fields for robot navigation and path planning in general high-dimensional domains, assuming the availability of a distance function that implicitly encodes the environment structure.

The main principle behind the method is that a range of potential fields for navigation can be encoded in the solution of the screened Poisson equation, which can be solved using the Walk on Spheres (WoS) method. WoS is a Monte Carlo method that is trivially parallelizable and scales well into higher dimensions, in contrast to the discretization and numerical method-based solvers generally used for lower dimensional environments.

In Section \ref{sec:empirical_eval} we empirically evaluate the characteristics of the WoS method in a simple environment for path planning applications and validate it in Section \ref{sec:validation_rr} an example planning problem using the RR platform. The results of the empirical evaluation and the validation studies in this paper motivate further research on potential improvements and applications of WoS-related methods for path planning and additional applications in robotics.

\bibliographystyle{unsrt}
\balance
\bibliography{references}

\end{document}